\documentclass[runningheads]{llncs}
\usepackage[T1]{fontenc}
\usepackage{graphicx}
\usepackage{cite}
\usepackage{booktabs}
\usepackage{multirow}
\usepackage{appendix}

\begin{document}
\title{Koopman-based surrogate modeling for reinforcement-learning-control of Rayleigh–Bénard convection}
\titlerunning{Koopman surrogate modeling for RL of Rayleigh–Bénard convection}
%
\author{Tim Plotzki \and Sebastian Peitz}
%
\authorrunning{Plotzki \& Peitz}
%
\institute{TU Dortmund University \& Lamarr Institute for Machine Learning and Artificial Intelligence, Dortmund, Germany}
%
\maketitle              
\begin{abstract}
Training reinforcement learning (RL) agents to control fluid dynamics systems is computationally expensive due to the high cost of direct numerical simulations (DNS) of the governing equations. Surrogate models offer a promising alternative by approximating the dynamics at a fraction of the computational cost, but their feasibility as training environments for RL is limited by distribution shifts, as policies induce state distributions not covered by the surrogate training data. In this work, we investigate the use of Linear Recurrent Autoencoder Networks (LRANs) for accelerating RL-based control of 2D Rayleigh-Bénard convection. We evaluate two training strategies: a surrogate trained on precomputed data generated with random actions, and a policy-aware surrogate trained iteratively using data collected from an evolving policy. Our results show that while surrogate-only training leads to reduced control performance, combining surrogates with DNS in a pretraining scheme recovers state-of-the-art performance while reducing training time by more than $40\%$. We demonstrate that policy-aware training mitigates the effects of distribution shift, enabling more accurate predictions in policy-relevant regions of the state space.

\end{abstract}
\section{Introduction}

Recent studies have shown that Reinforcement Learning (RL) is capable of controlling complex systems governed by fluid dynamics. 
Deep RL algorithms such as Proximal Policy Optimization (PPO) have been shown to produce state-of-the-art control strategies for fluid flows, and in particular Rayleigh-Bénard convection (RBC), which describes systems that are driven by buoyancy forces. These occur in a wide range of flavors, such as in the earth's atmosphere, in rooms with heated floors, or in nuclear fusion devices.
The dynamics are governed by nonlinear partial differential equations (PDEs) whose state is a function of both space and time, rendering direct numerical simulations (DNS) costly. Multi-query tasks such as uncertainty quantification, optimization or control are thus either limited to small problem setups, or require massive computational resources. 
At the same time, surrogate modeling has emerged as a computationally efficient alternative to DNS, capable of predicting the evolution of dynamical systems with reasonable accuracy, but existing research has largely focused on prediction rather than closed-loop control \cite{Pandey2022_datadrivenforecast,Markmann.2024,Fromme2025}. 

Rayleigh-Bénard convection describes the dynamics of a fluid subject to a vertical temperature difference \cite{Ahlers.2009,Pandey2018}. With increasing driving forces, the flow becomes increasingly turbulent and more difficult to control,
making it
an ideal testbed for studying feedback control of large distributed systems \cite{Becktepe2026afcbenchmark}.

A key challenge in surrogate modeling for RL is the \emph{distribution shift}. As the policy improves during training, the agent explores regions of the state space that differ from states typically seen in an uncontrolled RBC system. As a result, surrogates can become unreliable as they are asked to predict in a region of the state space not covered by the training data, limiting their usefulness for policy optimization and introducing performance-degrading biases.

This work addresses three key questions: (1) Can surrogate models provide sufficient accuracy to replace costly DNS during agent training? (2) Is it possible to combine surrogate predictions with DNS to reduce the agents training time without degrading control performance? (3) How can the distribution shift be mitigated to maintain surrogate reliability as the policy explores new regions of the state space?
Using the example of RBC, we perform numerical experiments to address these challenges, and give recommendations on how to intertwine surrogate modeling and reinforcement learning.

\section{Related Work}

RBC serves as a benchmark for surrogate modeling in fluid dynamics, particularly for evaluating data-driven reduced-order models under increasingly turbulent conditions \cite{Markmann2026FNO}. 
The linear recurrent autoencoder network (LRAN) has emerged as a popular architectural choice for nonlinear dynamical systems \cite{Otto.2019}. LRANs approximate the underlying Koopman operator \cite{Brunton2022} of a system by learning a nonlinear autoencoder and linear recurrent dynamics in the latent space simultaneously. They have been shown to produce good quantitative and qualitative predictions for the RBC system at a comparatively low computational cost \cite{Markmann.2024}.
At the same time, RBC has also been used for testing deep RL methods aimed at controlling complex dynamical systems \cite{Markmann.2025,Vignon.2023,Vasanth2024}. Agents trained on high-fidelity DNS using PPO have been shown to outperform traditional control strategies, 
and have been shown to generalize across a wide range of states. 
Besides, RL has been applied to a wide range of fluid systems such as channel flows \cite{Guastoni2023rlchannel}, decaying turbulence \cite{PSC+24}, aerodynamics \cite{RKJ+19} or the Kuramoto-Sivashinski equation \cite{Bucci2019}.

The area of model-based RL has become increasingly important in recent years, examples being the PILCO \cite{deisenroth2011pilco} or dreamer \cite{Hafner2019dreamer} algorithms as well as many recent developments on latent dynamics \cite{schwarzer2021dataefficient}.
Surrogate-based RL for PDEs has been investigated using deep learning (e.g., LSTMs \cite{WP24}) and neural operators \cite{Zhao2025}. A control framework combining data-driven manifold dynamics with RL (DManD-RL) \cite{Chen.2026} has been successful in controlling the two-dimensional RBC system by learning a policy inside a latent space.
The Koopman-operator framework has been applied to RL in various flavors, mainly to introduce linearity through lifting \cite{mondal2024efficient,rozwood2024koopman}. 

While Koopman-based surrogate modeling and deep RL have proven effective on the RBC system in isolation, this work investigates whether they can be combined to efficiently train RL agents. We train surrogates that predict the full state that allow for cheap rollouts while maintaining full compatibility with DNS-based control frameworks. 
We evaluate two training strategies for these surrogates. First, we train models on a static pre-generated dataset obtained from DNS. Second, we use a policy-aware training procedure that adapts the surrogate to policy-induced state distributions in order to mitigate distribution shift and produce models better suited for RL training.

\section{Methods}

\subsection{Rayleigh-Bénard Convection}

In RBC, a fluid between two plates is heated from below and cooled from above, creating buoyancy forces. Once the thermal forcing exceeds a critical threshold, buoyancy-driven convection sets in, observable as organized flow structures called \emph{convection cells} \cite{Ahlers.2009}.
The dynamics are governed by a system of partial differential equations derived from the incompressible Navier-Stokes equations. The state $\mathbf{s}_t = (T, u, w)$ at time $t$ is described by the temperature and velocity fields. 
We use standard fixed-temperature and no-slip boundary conditions (BCs) at the top and bottom boundaries and periodic BCs along the horizontal direction.

The system is characterized by two key quantities. The Rayleigh number $Ra$ measures the ratio of buoyancy-driven forces to dissipative effects caused by viscosity and thermal diffusion, where larger values lead to increasingly complex flow structures. 
We here focus on the moderately convective regime at $Ra = 10^4$. The second quantity is the Nusselt number $Nu$ quantifying the enhancement of heat transport due to convection relative to pure conduction
\footnote{$T_b, T_t$: (mean) temperature at bottom and top boundary, $\theta$: nondimensionalized temperature, $\kappa$: thermal diffusivity, $z$: vertical spatial coordinate, $d$: system height}:

$$Nu = \frac{\langle w\,\theta \rangle - \kappa\,\left\langle \frac{\partial \theta}{\partial z}\right\rangle}{\kappa\,(T_b - T_t) / d},$$
where $\langle \cdot \rangle$ denotes a spatial average. 
$Nu = 1$ corresponds to purely conductive heat transfer, while larger values indicate progressively stronger convective transport and flow complexity \cite{Ahlers.2009}.

Direct numerical simulation (DNS) is performed via the \emph{Oceananigans.jl} package \cite{Silvestri.2024} using a grid resolution of $96 \times 64$. To avoid transient effects, all experiments are initialized from precomputed checkpoints where the system is already in its convective phase. Separate sets of checkpoints are generated in advance for training, validation, and testing.

\subsection{Control of RBC}

Similar to \cite{Markmann.2025, Vignon.2023}, we focus on reducing $Nu$ by controlling $12$ thermal actuators located at the bottom boundary of the domain. Each actuator can locally set a temperature $T_i \in [1.25,\,2.75]$, subject to the constraint that their mean satisfies $\langle T \rangle = 2$. The actuators are driven by a normalized control signal $a_i \in [-1,1]$, which is mapped to the corresponding temperature range \cite{Markmann.2025}:
\begin{equation}
    \hat T_i' = a_i - \frac{\sum_{i=1}^N a_i}{N}, \quad 
    \hat T_i = \frac{0.75~\hat T_i'}{\max(1, |\hat T'|)}.
\label{eq:action_transform}
\end{equation}
The time between successive control inputs of the RL agent is set to $1.5$ seconds.

The policy is optimized using Proximal Policy Optimization (PPO)\cite{Schulman.2017}. PPO uses a clipped objective function that prevents large, potentially destabilizing policy updates. For model-free, continuous RL problems such as RBC, PPO has emerged as a widely used method due to its stability and robustness \cite{Markmann.2025}.
The policy is approximated by a multilayer perceptron, taking in coarsened temperature and velocity fields ($48 \times 8$, then flattened). It has a hidden layer with $64$ neurons, followed by the action values as outputs. ReLU is used as the activation function. The agent's reward signal is
$
R(\mathbf{s}_t) = 1 - \frac{Nu(\mathbf{s}_t)}{Nu_{\mathrm{base}}(Ra)}
$,
which corresponds to the negative Nusselt number normalized to the interval $[0,1]$. $Nu_{\mathrm{base}}(Ra)$ denotes the maximum Nusselt number. 

\subsection{Surrogate Architecture}

We extend the Linear Recurrent Autoencoder Network (LRAN) architecture, which has previously demonstrated strong predictive performance \cite{Otto.2019, Markmann.2024}, to the control context. LRAN learns a latent space in which the encoded variables can be advanced linearly in time. The encoder and decoder are convolutional neural networks (CNNs) with five and six layers, respectively. 
\begin{figure}[htbp]
    \centering
    \includegraphics[width=0.9\textwidth]{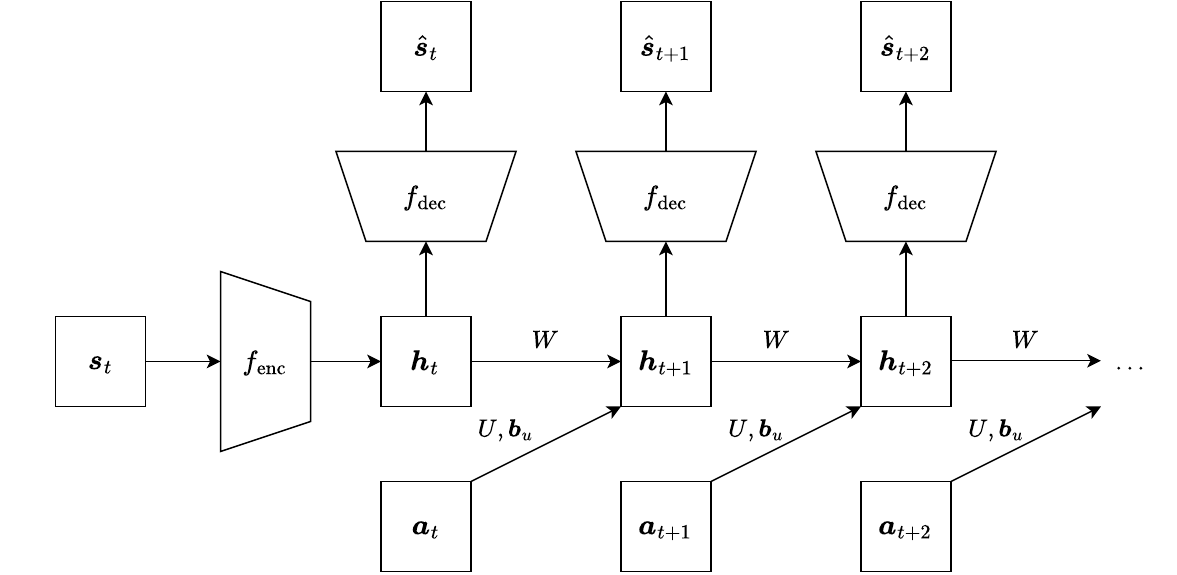}
    \caption{Extension of the LRAN architecture to incorporate control actions as additional inputs. The action vector $\mathbf{a}_t$ is added to the next hidden state after an affine transformation defined by an input-to-hidden matrix $U$ and a bias term $\mathbf{b}_u$. 
    }
    \label{fig:lran}
\end{figure}
%
The primary modification is the inclusion of control actions at each hidden state, as illustrated in Figure \ref{fig:lran}.
The model is optimized with respect to the loss 
\[
\mathcal{L} = \sum_{i = 0}^{\mathcal{T} - 1} \frac{\delta^{i}}{\mathcal{T}} \frac{|| \hat s_{t+i}-s_{t+i}||^2}{|| s_{t+i}||^2 + \epsilon},
\]
corresponding to a normalized reconstruction error and encourages accurate prediction of all observed states within a temporal sequence of length $\mathcal{T}$. The loss function introduces two additional hyperparameters: the sequence length $\mathcal{T}$ and a temporal discount factor $\delta$, which progressively decays reconstruction errors at later time steps in the sequence. The architecture is optimized with Adam \cite{Kingma.2017} with L2 regularization $\lambda = 10^{-4}$, using gradients accumulated over an entire sequence of length $\mathcal{T}$.

\subsubsection{Training}

\begin{figure}[htbp]
    \centering
    \hspace*{-1.5cm}
    \includegraphics[width=0.6\textwidth]{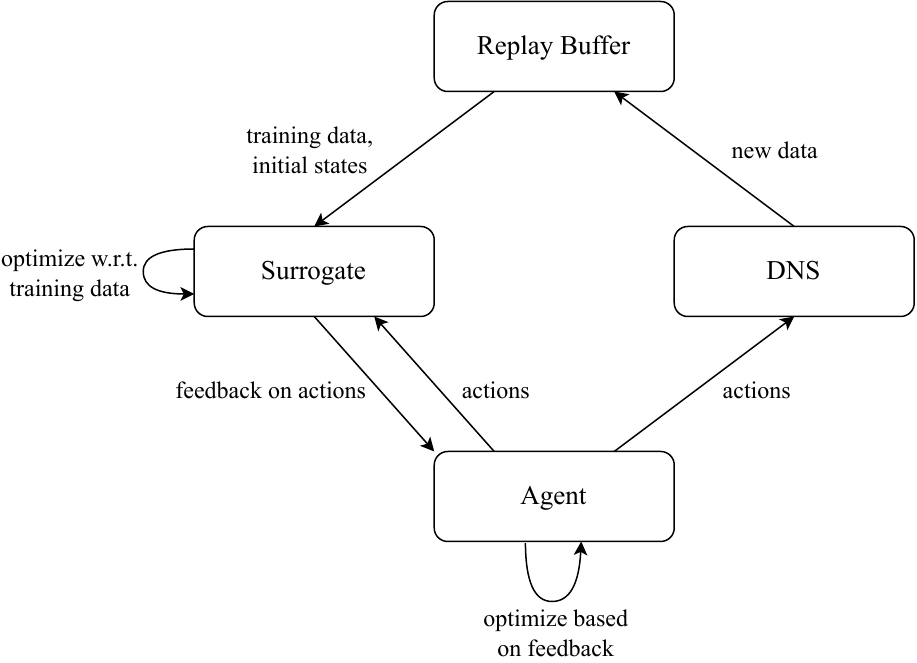}
    \caption{Training loop of the policy-aware surrogate training scheme. The surrogate model is optimized with data from DNS using actions from the policy while the policy is optimized by interacting with the surrogate using PPO.}
    \label{fig:hybrid_scheme}
\end{figure}

After a hyperparameter search, the configuration chosen for training the LRANs is $(\dim(h)=200, \delta=0.9, \mathcal{T}=10)$, where $\dim(h)$ refers to the dimensionality of the latent space. Furthermore, a learning rate of $\alpha = 5 \cdot 10^{-5}$ is used. Two LRANs are trained (see Fig.\ \ref{tab:lran_architecture} for details): one using a precomputed dataset with $3300$ episodes containing $400$ steps each using random actions, and one using a policy-aware training scheme inspired by Model-Based Policy Optimization (MBPO) \cite{Janner.2021}, in which training data is generated on the fly using actions from a policy that is optimized alongside the surrogate. Figure \ref{fig:hybrid_scheme} illustrates the training loop of this policy-aware training scheme.

\subsubsection{Data Augmentation}

Since data generation using DNS is computationally expensive, data efficiency is improved by augmenting existing episodes using translation and reflection. Because of the periodic boundary conditions along the horizontal direction, translation preserves the RBC dynamics if grid points shifted out of the domain are wrapped around to the opposite side. In our setup, with a horizontal grid resolution of $96$ and $12$ thermal actuators, the domain can be partitioned into segments of eight grid points, corresponding to the width of a single actuator. This allows for discrete translations by segments, yielding $12$ distinct configurations. In addition, each configuration can be reflected about the vertical axis, resulting in a total of $23$ distinct synthetic episodes per simulated episode.

\section{Experiments and Results}

We test both LRANs in two separate experiments to evaluate their suitability as training environments for RL control. First, we measure the control performance and training times of agents trained exclusively on the surrogates. Second, we employ the surrogates in a pretraining scheme in which training initially starts on the surrogate before finishing in a DNS environment. The idea is to start learning on the computationally efficient surrogate and only use expensive DNS to finetune control performance once surrogate imperfections prevent further progress. To accurately determine an agent's capabilities, the reported control performance is the mean Nusselt number measured in $20$ DNS environments initialized from random test checkpoints.

The PPO setup is largely identical to that used in \cite{Markmann.2025}. The algorithm is implemented using the \emph{Stable-Baselines3} library \cite{Raffin.2021}. We use $20$ parallel environments, an episode length of $t = 200$, a clipping range of $\epsilon = 0.2$, a discount factor of $\gamma = 0.99$, a learning rate of $\alpha = 10^{-3}$, and an entropy loss coefficient of $\beta = 0.01$. All other parameters are kept at their default values. Experiments are conducted at a Rayleigh number of $Ra = 10^4$. Time measurements are from $20$ cores of an \emph{Intel Xeon E5-2690v4} CPU together with an \emph{NVIDIA Tesla P100} GPU.

\begin{table}[t]
\caption{Encoder and decoder architectures of the LRAN. Here, $B$ denotes the batch size and $\dim(h)$ the latent dimension. Layers correspond to individual PyTorch classes. A Gaussian Error Linear Unit (GELU) \cite{Hendrycks.2023} is used as the activation function.}
\centering
\small
\renewcommand{\arraystretch}{1.15}
\begin{tabular}{|l|c||l|c|}
\hline
\multicolumn{2}{|c||}{\textbf{Encoder}} & \multicolumn{2}{c|}{\textbf{Decoder}} \\
\hline
Layer & Output shape & Layer & Output shape \\
\hline
Input & $[B, 3, 64, 96]$ &
Input & $[B, \dim(h)]$ \\
Conv2d $(3 \rightarrow 32,\;5\times5)$ & $[B, 32, 64, 96]$ &
Linear & $[B, 12288]$ \\
MaxPool $2\times2$ & $[B, 32, 32, 48]$ &
Unflatten & $[B, 32, 16, 24]$ \\
Conv2d $(32 \rightarrow 64,\;5\times5)$ & $[B, 64, 32, 48]$ &
Conv2d $(32 \rightarrow 64,\;5\times5)$ & $[B, 64, 16, 24]$ \\
Conv2d $(64 \rightarrow 32,\;5\times5)$ & $[B, 32, 32, 48]$ &
Upsample $\times2$ & $[B, 64, 32, 48]$ \\
MaxPool $2\times2$ & $[B, 32, 16, 24]$ &
Conv2d $(64 \rightarrow 32,\;5\times5)$ & $[B, 32, 32, 48]$ \\
Conv2d $(32 \rightarrow 32,\;5\times5)$ & $[B, 32, 16, 24]$ &
Conv2d $(32 \rightarrow 32,\;3\times3)$ & $[B, 32, 32, 48]$ \\
Flatten & $[B, 12288]$ &
Upsample $\times2$ & $[B, 32, 64, 96]$ \\
Linear & $[B, \dim(h)]$ &
Conv2d $(32 \rightarrow 16,\;3\times3)$ & $[B, 16, 64, 96]$ \\
 &  &
Conv2d $(16 \rightarrow 3,\;5\times5)$ & $[B, 3, 64, 96]$ \\
\hline
\end{tabular}
\label{tab:lran_architecture}
\end{table}

\subsection{Experiment 1: Exclusive Training on Surrogates}

We evaluate the control performance of agents after being trained in different environments. Training lasts until performance stagnates and no more progress is made. Agents are compared to two baselines: "Zero" denotes a policy that always outputs $a_t = \{0\}^{12}$, which is equivalent to letting the system evolve without control. The "Random" baseline corresponds to a policy that samples actions uniformly from the action space $a_t \sim \mathcal{U}([-1,1]^{12})$ at every time step. Additionally, we consider the control performance and training time of an agent trained directly in a DNS environment using $400{,}000$ interactions, following the experimental setup of \cite{Markmann.2025}. The results are summarized in Table \ref{tab:results1}.

\begin{table}[htbp]
    \caption{Control performance measured by the Nusselt number $\mathrm{Nu}$ and total training time for policies from different environments.}
    \setlength{\tabcolsep}{8pt}
    \centering
    \begin{tabular}{lrr}
        \toprule
        \textbf{Environment} & \textbf{Nu} & \textbf{Training time} \\
        \midrule
        Random-Action & \multirow{2}{*}{$3.31$} & \multirow{2}{*}{\textbf{$\mathbf{0}$~h $\mathbf{06}$~min}} \\
        $200{,}000$ interactions & & \\[3pt]
        Policy-Aware & \multirow{2}{*}{$2.97$} & \multirow{2}{*}{$0$~h $17$~min} \\
        $600{,}000$ interactions & & \\[3pt]
        DNS & \multirow{2}{*}{$\mathbf{2.74}$} & \multirow{2}{*}{$4$~h $11$~min} \\
        $400{,}000$ interactions & & \\
        \midrule
        Zero & $4.00$ & - \\
        Random & $4.05$ & - \\
        \bottomrule
    \end{tabular}
    \label{tab:results1}
\end{table}

All surrogate environments produce agents that significantly outperform both baselines. However, their control performance still remains below that of the DNS-trained agent. Because surrogate rollouts are on average $25.6$ times faster than DNS simulations, the total training time is substantially reduced even when more interactions are performed.

\begin{figure}[htbp]
    \centering
    \includegraphics[width=0.65\textwidth]{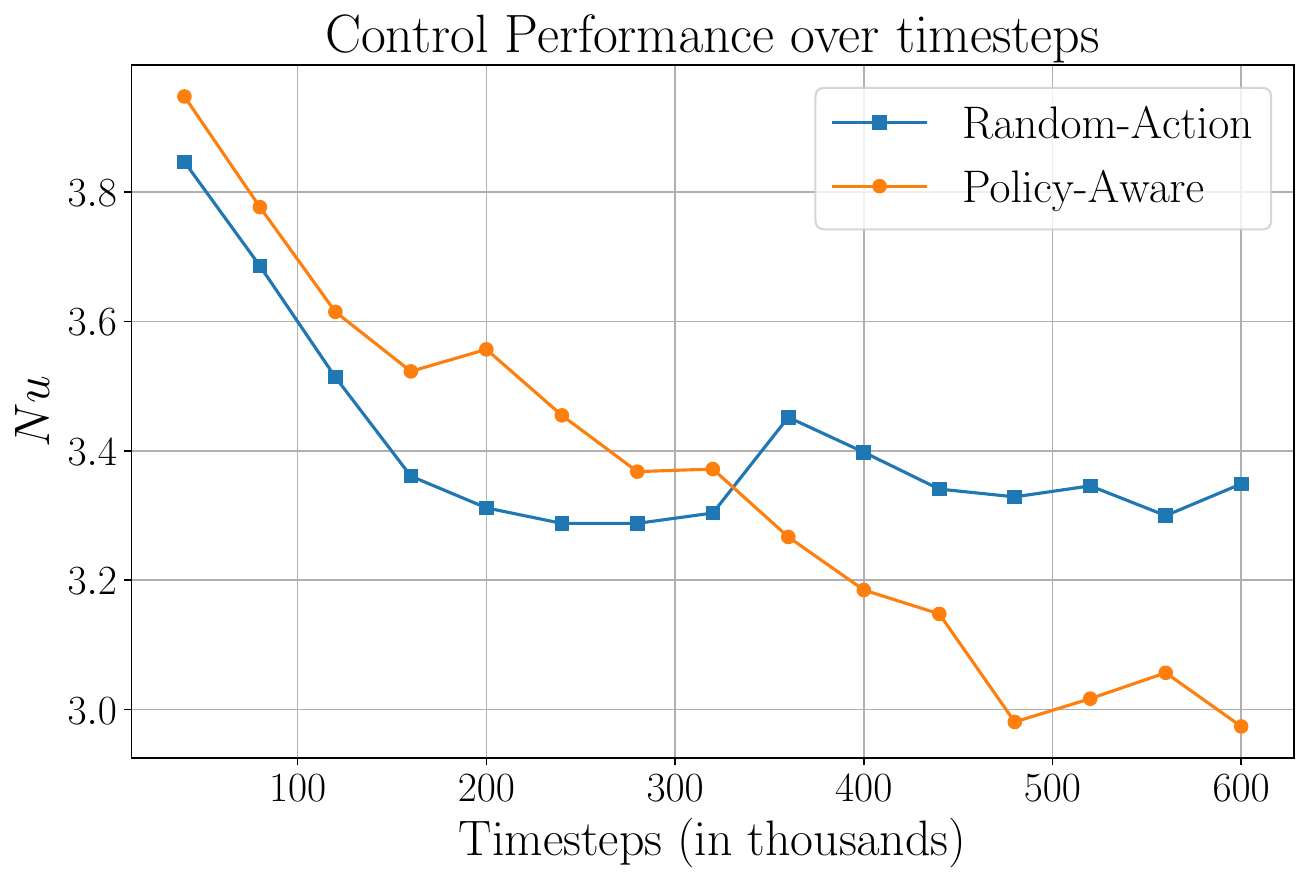}
    \caption{Control performance of policies from both surrogates over the course of training.}
    \label{fig:performance_timesteps}
\end{figure}

The Random-Action surrogate quickly reaches a control performance of $Nu = 3.31$ after only $200{,}000$ interactions, after which learning plateaus. Additional PPO updates fail to further improve the agent's true performance when evaluated in DNS. In contrast, the Policy-Aware surrogate exhibits slower initial learning but improves steadily throughout training. After approximately $350{,}000$ interactions it surpasses the Random-Action surrogate and continues to improve thereafter. Using the Policy-Aware surrogate as the training environment, the agent eventually reaches a control performance of $Nu = 2.97$ after $600{,}000$ interactions, corresponding to a total training time of $17$ minutes. Figure \ref{fig:performance_timesteps} compares policy performance during training on the surrogates.

\subsection{Experiment 2: Pretraining on the Surrogates}

In this experiment, the surrogates are used in a pretraining scheme in which training starts in a surrogate environment before continuing in DNS. The goal is to exploit the rapid early learning enabled by the surrogate while relying on DNS to fine-tune the policy once surrogate inaccuracies begin to limit further improvements. Table \ref{tab:results2} summarizes the results.

\begin{table}[htbp]
    \caption{Control performance and total training time for policies from two different pretraining configurations compared to a DNS environment.}
    \setlength{\tabcolsep}{8pt}
    \centering
    \begin{tabular}{lrr}
        \toprule
        \textbf{Environment} & \textbf{Nu} & \textbf{Training time} \\
        \midrule
        Random-Action + DNS & \multirow{2}{*}{$\mathbf{2.73}$} & \multirow{2}{*}{$3$~h $06$~min} \\
        $120{,}000$ + $280{,}000$ interactions & & \\[3pt]
        Policy-Aware + DNS & \multirow{2}{*}{$2.75$} & \multirow{2}{*}{\textbf{$\mathbf{2}$~h $\mathbf{24}$~min}} \\
        $400{,}000$ + $200{,}000$ interactions & & \\[3pt]
        DNS & \multirow{2}{*}{$2.74$} & \multirow{2}{*}{$4$~h $11$~min} \\
        $400{,}000$ interactions & & \\
        \bottomrule
    \end{tabular}
    \label{tab:results2}
\end{table}

Both surrogates are capable of reaching the control performance of a purely DNS-trained agent in a pretraining scheme while reducing the overall training time. As in the previous experiment, the Random-Action surrogate provides faster performance gains during the early stages of training. This allows the agent to achieve a slightly improved $Nu$ within the same total number of interactions, of which $30\%$ are performed on the surrogate model.

The Policy-Aware surrogate, however, requires more total interaction steps to compensate for slower early learning. Nevertheless, the stronger policies eventually obtained when training on this surrogate allow the number of required DNS interactions to be reduced further. As a result, a similar control performance is achieved with an even lower total training time.

\section{Discussion}

Our results demonstrate that surrogate models, specifically the LRAN, are capable of producing agents significantly better than baseline policies. In conjunction with DNS using a pretraining scheme, surrogates allow policies to match state-of-the-art control performance while requiring fewer DNS interactions and consequently less total training time.

\begin{figure}[htbp]
    \centering
    \includegraphics[width=\textwidth]{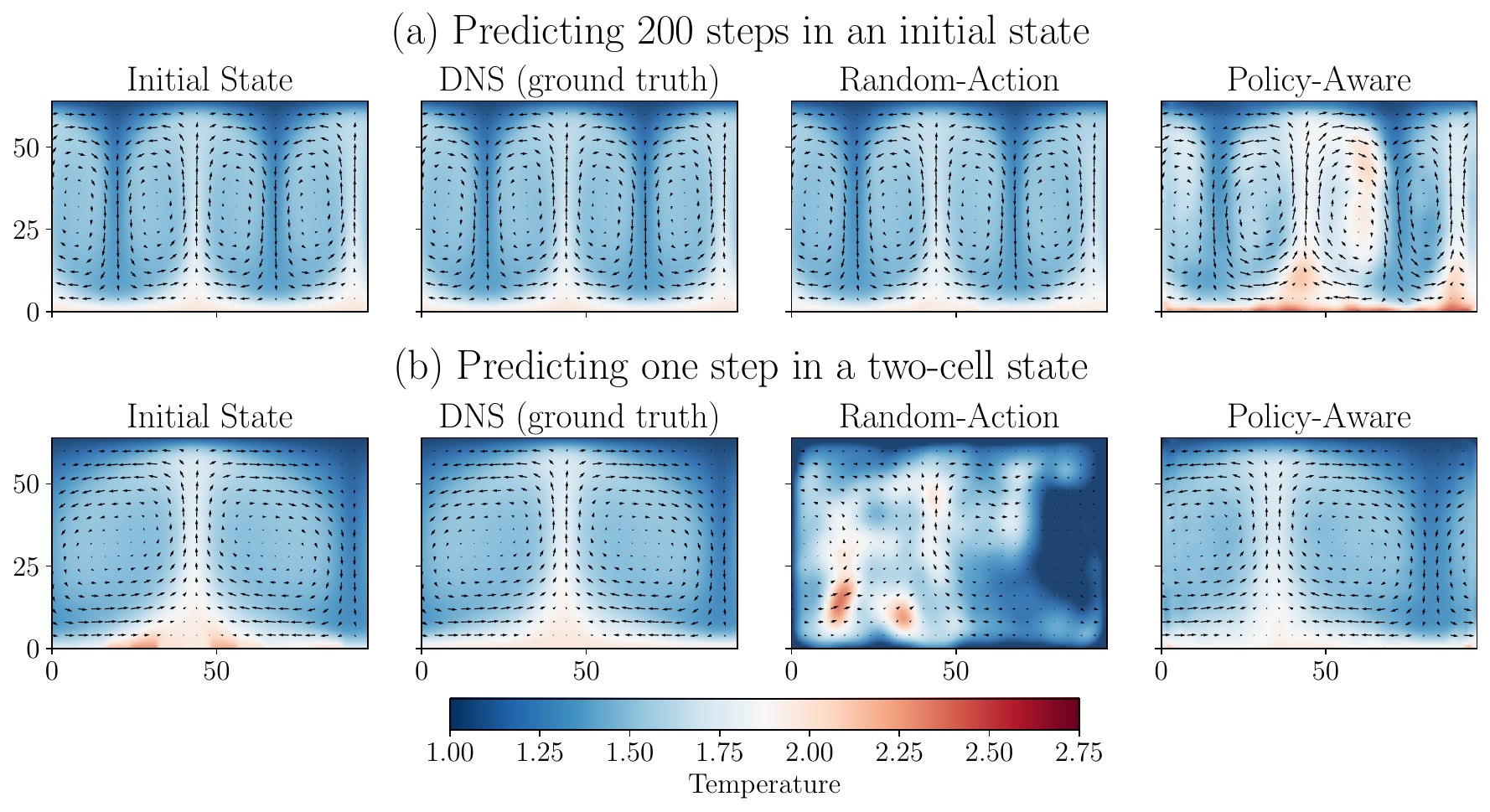}
    \caption{Qualitative comparison of both LRANs. (a) Surrogates predict $200$ steps with zero action input in an initial four-cell RBC state. (b) Surrogates predict one step with zero action input in a two-cell RBC state. Control with zero action inputs is equivalent to uncontrolled forward prediction.}
    \label{fig:predictions}
\end{figure}

Interestingly, the two LRANs demonstrate different strengths. All agents discover the \emph{cell-merging} strategy during training, which has already been observed in \cite{Markmann.2025}. This strategy collapses the initial four-cell RBC state into two larger cells, ultimately reducing convection and providing a visible example of the distribution shift. However, since the static dataset used for the Random-Action surrogate is generated from random actions that follow no strategy, such states are severely underrepresented. As a consequence, the Random-Action surrogate is unable to predict two-cell states. In contrast, the Policy-Aware model has learned to predict two-cell RBC states, but its accuracy degrades in the initial four-cell state space, where it overestimates the chaotic behavior caused by unoptimized zero-input actions. The Random-Action surrogate does not display this issue to the same degree. A plausible explanation is that the Policy-Aware surrogate overfits to optimized actions, since its training data samples actions from a policy trained alongside the model. These effects are illustrated in Figure \ref{fig:predictions}.

This observation explains the training behavior of agents in the previous experiments. Policies optimized with the Policy-Aware surrogate exhibit slower initial learning due to the reduced model accuracy in the initial four-cell state space. However, once two-cell states become relevant for further optimization, the Policy-Aware model begins to outperform the Random-Action surrogate, whose training plateaus because it cannot predict these states.

These findings suggest that surrogate models used as RL training environments should account for policy-induced state distributions during training to mitigate the effects of distribution shift. Otherwise, the surrogate may fail to predict states that emerge as the policy improves and therefore lose reliability as an RL environment. However, the extent to which policy-aware training is necessary will likely depend on the characteristics of the underlying system and the degree to which policy optimization alters the relevant state distribution.

\section{Conclusion}

This work showed that surrogate models can be used as environments for training RL agents. We examined different training approaches for the surrogates, highlighted the impact of distribution shift on downstream policy performance, and presented an approach to mitigate it. Furthermore, we demonstrated that surrogates can be combined with DNS in a pretraining scheme to match state-of-the-art control performance while reducing agent training time.

While this study focused on 2D RBC with a moderate Rayleigh number, future work could investigate the LRAN's ability to generalize to higher Rayleigh numbers or to different dynamical systems altogether. Furthermore, although this work trained surrogates better suited for RL using a policy-aware training approach, future work could explore alternative policy-aware schemes that reduce the overfitting problem observed here. This may further reduce training times and lead to surrogate models capable of producing even stronger control policies.

\begin{credits}


\subsubsection{\ackname} 
SP acknowledges funding from the European Research Council (ERC Starting Grant ``KoOpeRaDE'') under the European Union’s Horizon 2020 research and innovation programme (Grant agreement No. 101161457).
We gratefully acknowledge the computing time provided on the Linux HPC cluster at Technical University Dortmund (LiDO3), partially funded in the course of the Large-Scale Equipment Initiative by the German Research Foundation (DFG, Deutsche Forschungsgemeinschaft) as project 271512359.

\end{credits}
%
%
%
\bibliographystyle{splncs04}
\bibliography{literatur}

\end{document}